%% file: Yousef2008Multinormal.tex
\def\MYJOURNAL{0} 

\if 0\MYJOURNAL%
\documentclass[3p, preprint, authoryear]{elsarticle}
\usepackage{setspace}
\makeatletter
\def\ps@pprintTitle{%
  \let\@oddhead\@empty
  \let\@evenhead\@empty
  \def\@oddfoot{\reset@font\hfil\thepage\hfil}
  \let\@evenfoot\@oddfoot
}
\makeatother
\else \if 1\MYJOURNAL%
  \documentclass[journal,10pt,twocolumns,letter]{IEEEtran}
  \usepackage[numbers,sort&compress]{natbib}
\fi
\fi

\usepackage{geometry}%
\usepackage{color, xcolor, colortbl}
\usepackage{hyperref}
\hypersetup{
  colorlinks,
  citecolor=blue,
  filecolor=blue,
  linkcolor=blue,
  urlcolor=blue
}
\usepackage{amsfonts}
\usepackage{amsmath,bm}
\usepackage{amssymb}
\usepackage{graphicx}
\usepackage{url}
\usepackage{fp}
\usepackage{subfig}

\providecommand{\U}[1]{\protect\rule{.1in}{.1in}}

\usepackage{enumitem}
\usepackage[referable,para]{threeparttablex}
\usepackage{footnote}
\usepackage{booktabs}  

\usepackage{mwe}
\usepackage[percent]{overpic}
\usepackage{listings}
\definecolor{dkgreen}{rgb}{0,.6,0}
\definecolor{dkblue}{rgb}{0,0,.6}
\definecolor{dkyellow}{cmyk}{0,0,.8,.3}

\lstdefinestyle{customphp}{
  language        = php,
  basicstyle      = \small\ttfamily,
  keywordstyle    = \color{dkblue},
  stringstyle     = \color{red},
  identifierstyle = \color{dkgreen},
  commentstyle    = \color{gray},
  emph            =[1]{php},
  emphstyle       =[1]\color{black},
  emph            =[2]{if,and,or,else},
  emphstyle       =[2]\color{dkyellow}}

\lstdefinestyle{customc}{
  breaklines=true, breakindent=20pt,
  frame=leftline,
  numbers=left,
  language=C, numberstyle=\tiny, numbersep=10pt,
  showstringspaces=false,
  basicstyle=\footnotesize\ttfamily,
  keywordstyle=\bfseries\color{green!40!black},
  commentstyle=\itshape\color{purple!40!black},
  identifierstyle=\color{blue},
  stringstyle=\color{orange},
  captionpos=t
}

\graphicspath{{./Graphics/}}

\usepackage{fourier}
\DeclareMathAlphabet{\mathcal}{OT1}{pzc}{m}{it}
\DeclareSymbolFont{letters}{OML}{cmm}{m}{it}

\usepackage{tikz}
\usepackage{float}
\usepackage{pgf}
\usepackage{pgfplots}

\usetikzlibrary{calc}
\usepgflibrary{shapes}
\usetikzlibrary{er}
\usetikzlibrary{arrows,snakes,backgrounds}
\usetikzlibrary{decorations.text}

\usepgfplotslibrary{external}
\usetikzlibrary{spy}

\def\getangle(#1) (#2)#3{%
  \begingroup%
  \pgftransformreset%
  \pgfmathanglebetweenpoints{\pgfpointanchor{#1}{center}}{\pgfpointanchor{#2}{center}}%
  \expandafter\xdef\csname angle#3\endcsname{\pgfmathresult}%
  \endgroup%
}

\pgfplotsset{compat=1.11}
\usetikzlibrary{calc,trees,positioning,arrows,chains,shapes.geometric,%
  decorations.pathreplacing,decorations.pathmorphing,shapes,%
  matrix,shapes.symbols}

\tikzset{
  >=stealth',
  punktchain/.style={
    font=\scriptsize,
    rectangle,
    rounded corners,
    draw=black, thick,
    text width=10em,
    minimum height=1em,
    text centered},
  line/.style={draw, thick, <-},
  element/.style={
    tape,
    top color=white,
    bottom color=blue!50!black!60!,
    minimum width=8em,
    draw=blue!40!black!90, very thick,
    text width=10em,
    minimum height=1em,
    text centered},
  every join/.style={->, thick,shorten >=1pt},
  decoration={brace},
  tuborg/.style={decorate},
  tubnode/.style={midway, right=2pt},
}

\if 0\MYJOURNAL%
\tikzset{
  PIXEL/.style={
    font=\fontsize{4}{3.6}\selectfont,
    text width=9em,
    minimum height=1em,
    text centered
  }
}
\else \if 1\MYJOURNAL%
  \tikzset{
    PIXEL/.style={
      font=\tiny,
      text width=8em,
      minimum height=3em,
      text centered
    }
  }
\fi
\fi

\usepackage{todonotes}
\usepackage{changebar}

\graphicspath{{./Graphics/}}

\begin{document}

\input{SecAbstract.tex}

\input{SecIntroduction.tex}

\input{SecLoglikelihood.tex}

\input{SecPerofrmance.tex}

\input{SecConclusion.tex}

\input{SecAcknowledgment.tex}

\bibliographystyle{elsarticle-harv}
\bibliography{publications,booksIhave}

\end{document}

%% file: SecAbstract.tex
\begin{frontmatter}
  \title{Prudence When Assuming Normality:\\an advice for machine learning
    practitioners\tnoteref{t1}}\tnotetext[t1]{This manuscript was initially composed in 2006 as part
    of a the author's Ph.D. dissertation. This paper is currently under consideration in Pattern
    Recognition Letters.}

  \author[WAY]{Waleed~A.~Yousef\corref{cor1}}
  \ead{wyousef@GWU.edu, wyousef@fci.helwan.edu.eg}
  \cortext[cor1]{Corresponding Author}

  \address[WAY]{Ph.D., Computer Science Department, Faculty of Computers and Information, Helwan
    University, Egypt.\\ Human Computer Interaction Laboratory (HCI Lab.), Egypt.}

  \begin{abstract}
    In a binary classification problem the feature vector (predictor) is the input to a scoring
    function that produces a decision value (score), which is compared to a particular chosen
    threshold to provide a final class prediction (output). Although the normal assumption of the
    scoring function is important in many applications, sometimes it is severely violated even under
    the simple multinormal assumption of the feature vector. This article proves this result
    mathematically with a counter example to provide an advice for practitioners to avoid blind
    assumptions of normality. On the other hand, the article provides a set of experiments that
    illustrate some of the expected and well-behaved results of the Area Under the ROC curve (AUC)
    under the multinormal assumption of the feature vector. Therefore, the message of the article is
    not to avoid the normal assumption of either the input feature vector or the output scoring
    function; however, a prudence is needed when adopting either of both.
  \end{abstract}

  \begin{keyword}
    Bayes Classifier\sep Multinormal Distribution\sep Central Limit Theorem\sep Classification\sep
    binormal model.
  \end{keyword}
\end{frontmatter}


%% file: SecIntroduction.tex
\section{Introduction}\label{sec:introduction}
\subsection{Formalization and Notation}\label{sec:form-notat}
\begin{figure}[t]\centering
  \includegraphics[height=2.5in]{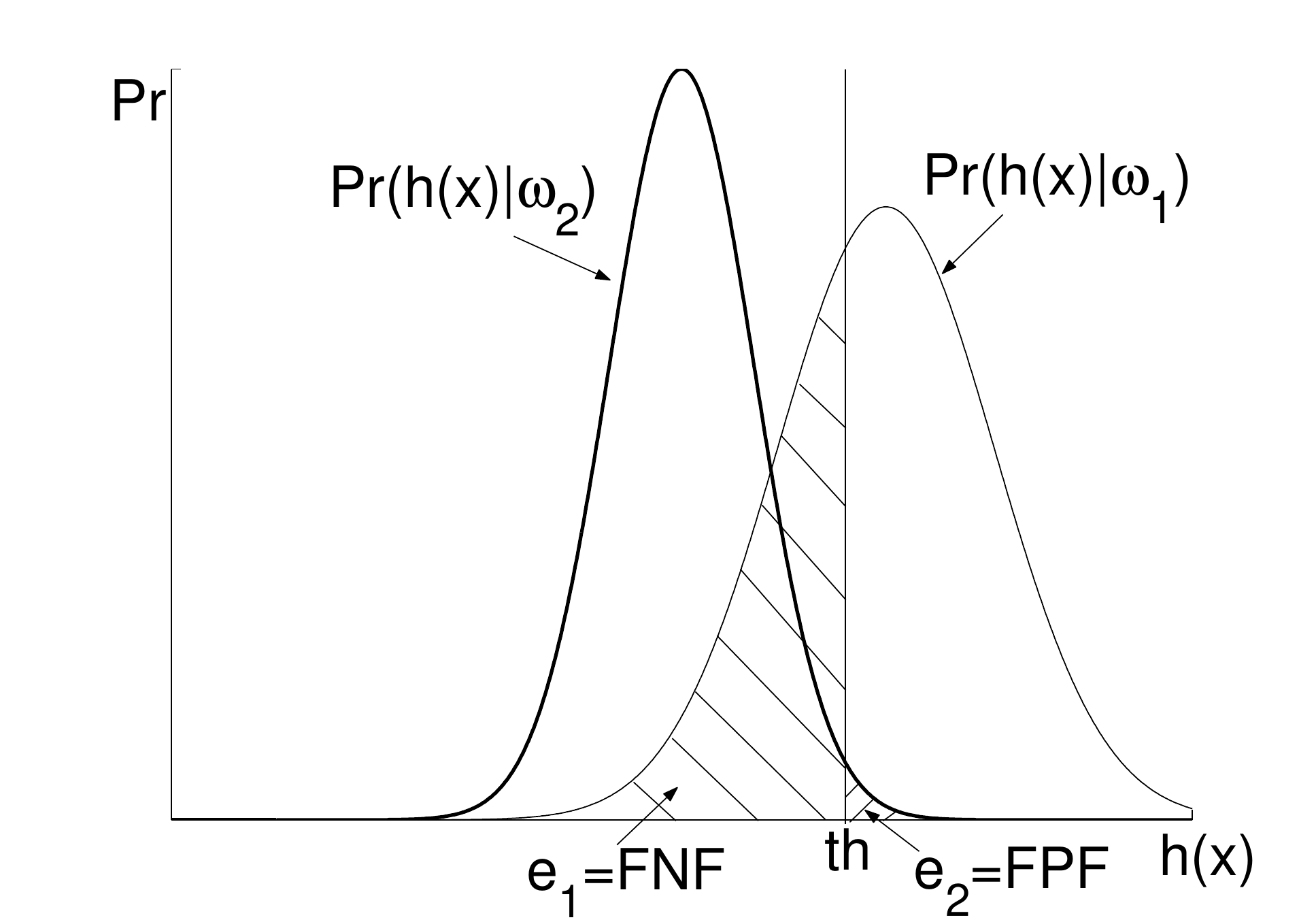}
  \caption{The probability of log-likelihood ratio conditional under each class. The two
    components of error are indicated as the FPF and FNF, the conventional terminology in medical
    imaging.}\label{fig3}%
\end{figure}

If a random vector $X$ and a random variable $Y$ have a joint probability density $f_{X,Y}(x,y)$, and
$Y$ is qualitative (or categorical) with only two possible values (class $\omega_1$ or class
$\omega_2$) the binary classification problem is defined as follows: how to find the classification
rule $\hat{Y} = \eta(X)$ that is able to predict the class $Y$ (the response) from an observed value
for the variable $X$ (the predictor). The prediction function is required to have minimum average
prediction error, which is defined in terms of some loss function $L(Y,\eta(X))$ that penalizes for
any deviation in the predicted value $\hat{Y}$ of the response from the correct value $Y$. Since $Y$
takes one of two values ($\omega_1$ or $\omega_2$), the loss function can be defined by the matrix
$ L(Y, \hat{Y})=((c_{ij})),\ i,j = 1,2 $, where the non-negative element $c_{ij}$ is the cost, the
penalty or the price, paid for classifying an observation as $\omega_{j}$ when it belongs to
$\omega_{i}$. Then, the risk of this prediction function is defined by the average loss, according
to the defined loss function, which can be written as $R(\eta)=E\left[ {L(Y,\hat{Y})}\right]$. These
foundations can be found in, e.g., \cite{Hastie2009ElemStat, Duda2001PatternClassification,
  Bishop2006PatRecMachInt}. It is known from the statistical decision theory that the rule $\eta$
that minimizes this risk $R$ is given by
\begin{equation}
  \eta(x):\quad \frac{f_{X}(X=x|\omega_{1})}{f_{X}(X=x|\omega_{2})}\underset{\omega_{2}}{\overset{\omega_{1}}{\gtrless}}\frac{\Pr\left[\omega_{2}\right]  \left({c_{22}-c_{21}}\right)}{\Pr\left[\omega_{1}\right]  \left(  {c_{11}-c_{12}}\right)  }.\label{eq14}
\end{equation}
However, in the feature subspace, $X \in \mathbf{R}^p$, the regions of classification (the solution
to the inequality \eqref{eq14}) have the dimensionality $p$, and it is very difficult to calculate
the error components from multi-dimensional integration. It is easier to look at (\ref{eq14}) as:%
\begin{subequations}\label{eq:OptimalRule}
  \begin{gather}
    \eta(x):\quad h(x)\underset{\omega_{2}}{\overset{\omega_{_{1}}}{\gtrless}}th,\label{eq45}\\
    h(x)  =\log\frac{f_{X}(X=x|\omega_{1})}{f_{X}(X=x|\omega_{2})},\label{eq:Logliklyhood}\\
    th =\log\frac{\Pr\left[ \omega_{1}\right] \left( {c_{22}-c_{21}}\right) }{\Pr\left[ \omega_{2}\right](c_{11}-c_{12})},\label{eq:threshold}
  \end{gather}%
\end{subequations}
and $h(X)$ is called the log-likelihood ratio and acts as the decision value or the score. Now the
log-likelihood ratio itself is a random variable whose variability comes from the feature vector
$X$, and has a PDF conditional on the true class. This is shown in \figurename~\ref{fig3}. It can be
easily shown that the two curves in \figurename~\ref{fig3} cross at $h(X)=0$, when the threshold is
zero ($th = 0$).

Now, if the joint distribution is known but its parameters are not known, a learning process is used
to estimate those parameters from a training sample $\mathbf{t}$ by methods of statistical
inference. Moreover, if the joint distribution is unknown, a classification function is modeled
parametrically or nonparametrically and a training sample is used to build that model. This is the
field of statistical learning, machine learning, or the recently so called ``data science''. In
either cases there is a modeled decision function (or scoring function) $h(X)$, that is no longer
the log-likelihood ratio, and the final classification rule takes the form~\eqref{eq45}. However, of
course, it is no longer the optimal classification rule that minimizes the risk.

\bigskip

Regardless of how we obtain the decision (scoring) function $h(X)$, there should be some measure to
assess the performance of the classification rule. Said differently, if several classifiers are
competing in the same problem, which is better? One natural answer is to define the two types of
error: $e_{1}$ is the probability of classifying a case as belonging to class 2 when it belongs to
class 1, and $e_{2}$ is vice versa. Formally, this is defined as:
\begin{subequations}
  \label{Eqe1e2}
  \begin{align}
    e_{1}  &  =\int_{-\infty}^{th}{f_{h}}\left(  {h(x)|\omega_{1}}\right){dh(x)},\\
    e_{2}  &  =\int_{th}^{\infty}{f_{h}}\left(  {h(x)|\omega_{2}}\right)  {dh(x)}
  \end{align}%
\end{subequations}%
It is conventional in some fields, e.g., medical imaging, to refer to $e_{1}$ as the False Negative
Fraction (FNF), and $e_{2}$ as the False Positive Fraction (FPF). This is because diseased patients
typically have a higher output value for a test than non-diseased patients. For example, a patient
belonging to class 1 whose test output value is less than the threshold setting for the test will be
called \textquotedblleft test negative\textquotedblright\ while the patient is in fact in the
diseased class. This is a false negative decision; hence the name FNF. The situation is reversed for
the other error component. These components are illustrated in \figurename~\ref{fig3}.

\bigskip

Now assume the classifier is trained under the condition of equal prevalence and cost, i.e.,
$th = 0$. In other environments there will be different a priori probabilities yielding to different
threshold values. The error is not a sufficient measure now, since it is function of a single fixed
threshold. A more general way to assess a classifier is provided by the Receiver Operating
Characteristic (ROC) curve. This is a plot for the two components of error, $e_{1}$ and $e_{2}$
under different threshold values. However, a convention in several fields, e.g., medical imaging, is
to plot the $TPF=1-FNF$ vs. the $FPF$.

\begin{figure}[t]\centering
  \includegraphics[height=2.5in]{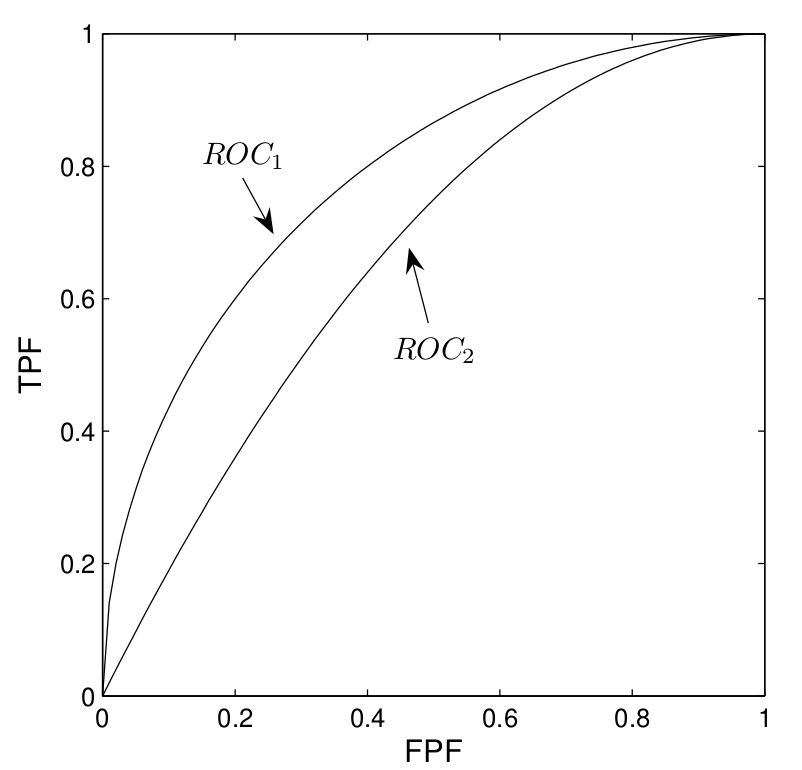}
  \caption{ROC curves for two different classifiers. ROC$_{1}$ is better than ROC$_{2}$, since for any
    error component value, the other component of classifier 1 is less than that one of classifier
    2.}\label{fig4}
\end{figure}

Since the error components~\eqref{Eqe1e2} are integrals over a particular PDF, the resulting ROC is
a monotonically non-decreasing function. In that case, the farther apart the two distributions of
the score function from each other, the higher the ROC curve and the larger the area under the curve
(AUC). \figurename~\ref{fig4} shows ROC curves for two different classifiers.
The first one performs better since it has a lower value of $e_{2}$ at each value of $e_{1}$. Thus,
the first classifier unambiguously separates the two classes better than the second one. Also, the
AUC for the first classifier is larger than that for the second one. AUC can be thought of as one
summary measure for the ROC curve. Formally the AUC is given by:
\begin{equation}
  AUC=\int_{0}^{1}{TPF~d(FPF)} \label{eq47}%
\end{equation}
If two ROC curves cross, this means each is better than the other for a certain range of the
threshold setting, but it is worse in another range. In that case some other measure can be used,
such as the partial area under the curve (PAUC) in a specified region, which is quite out of the
scope of the present article, e.g., see \cite{Jiang1996AReceiverOperating, Walter2005ThePartialArea,
  Yousef2013PAUC}. It is easy to show that the AUC defined in~\eqref{eq47} is equivalent to:%
\begin{align}
  AUC &=\Pr\left[h\left( {x|\omega_{2}}\right) < h\left(  {x|\omega_{1}}\right) \right],\label{eq106}
\end{align}
which expresses again the separation between the two sets of decision scores $h(X|\omega_1)$ and
$h(X|\omega_2)$.

\bigskip

The expressions~\eqref{Eqe1e2},~\eqref{eq47}, and \eqref{eq106} are population parameters that are
calculated from integrals assuming the full knowledge of distributions. However, in practical setups
only a finite testing dataset is available and only estimates of these values are available. The
Uniform Minimum Variance Unbiased Estimators (UMVUE) under the nonparametric assumptions are given
by \citep{Randles1979IntroductionTo,Hajek1999TheoryOfRank}:
\begin{subequations}\label{eq:2}
  \begin{align}
    \hat{e}_1 & = \sum\limits_{i=1}^{n_1} I_{h(x_{i}|\omega_{1})<th}\\
    \hat{e}_2 & = \sum\limits_{j=1}^{n_2} I_{h(x_{j}|\omega_{2})>th}\\
    \widehat{AUC}&=\frac{1}{n_{1}n_{2}}\sum\limits_{j=1}^{n_{2}}{\sum\limits_{i=1}^{n_{1}}{\psi\left(  {h\left(  {x_{i}|\omega_{1}}\right),h\left(  {x_{j}|\omega_{2}}\right)  }\right)  }},\label{eq58}\\
    \psi(a,b)&=\left\{%
               \begin{array}[c]{ccc}%
                 1 &  & a>b\\
                 1/2 &  & a=b\\
                 0 &  & a<b
               \end{array}\right.
  \end{align}
\end{subequations}
It is worth mentioning that, for a particular classification rule trained on the training data set
$\mathbf{t}$, the expressions~\eqref{Eqe1e2},~\eqref{eq47}, and \eqref{eq106} and their
estimates~\eqref{eq:2} all should be subscripted by $\mathbf{t}$ since everything will vary with
varying $\mathbf{t}$. We usually drop the subscript when there is no ambiguity.

\subsection{Background and Motivation}\label{sec:backgr-motiv}
From the introduction above, it is clear that a classifier trained on the training set $\mathbf{t}$
produces a decision (scoring) function $h_{\mathbf{t}}(X)$ that has its own distribution under each
class. This distribution determines the errors, ROC, AUC, and any other performance measure. It is
always tempting to assume normality for a random variable for easier mathematical analysis or
statistical inference. For binary classification problem, in particular, normality can be assumed
for either the final scoring function and/or the input feature vector. This will be elaborated next
in the following two paragraphs.

\bigskip

\begin{figure}[t]\centering
  \includegraphics[height=2in]{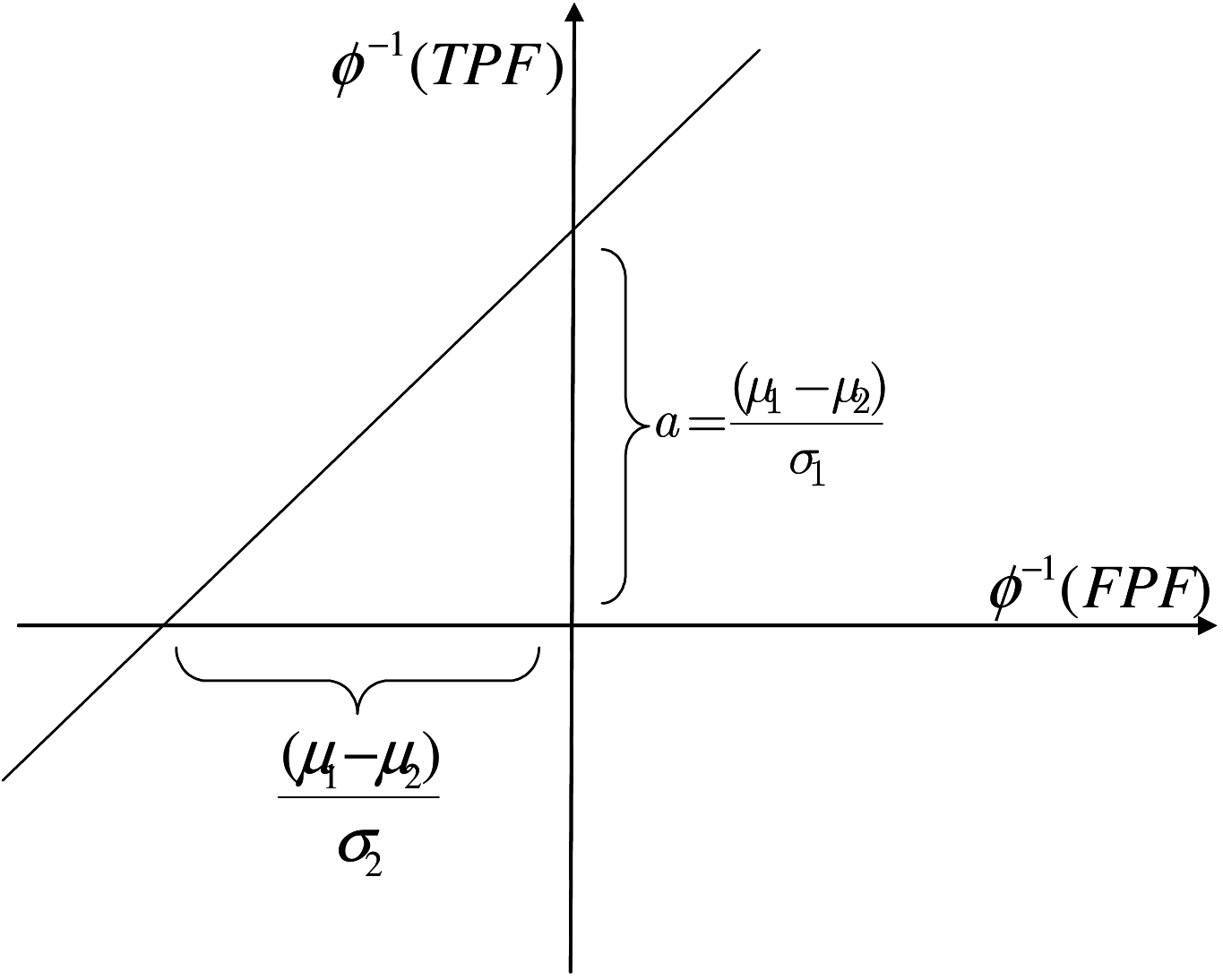}
  \caption{The double-normal-deviate plot for the ROC under the normal assumption for the
    log-likelihood ratio is a straight line.}\label{figNormalDeviate}
\end{figure}

First, the normality assumption of the two distributions of the scoring function $h(X)$ (under the
two classes) is known as binormal model and has very interesting properties. These properties, along
with their proofs exist in many texts and articles including very early literature, e.g.,
\cite{Green1966SignalDetectionTheoryPsychophysics, Dorfman1992ROCRating, PanMetz1997ProperROC,
  MetzPan1999ProperROC, Dorfman1997PropROCbigamma, Krzanowski2009ROC}. Under this assumption, the
errors~\eqref{Eqe1e2} clearly are given by a simple integration over the tail of normal
distributions; and interestingly, the ROC curve is expressed using the inverse error function
transformation, as:
\begin{equation}
  \phi^{-1}(TPF)=\frac{(\mu_{1}-\mu_{2})}{\sigma_{1}}+(\frac{\sigma_{2}}{\sigma_{1}})\phi^{-1}(FPF).\label{eq:NormalDeviate}
\end{equation}
This means that the whole ROC curve can be summarized in just two parameters: the intercept, and the
slope; this is shown in \figurename~\ref{figNormalDeviate}. Denote the intercept and slope
of~\eqref{eq:NormalDeviate} by $a$ and $b$ respectively; it is easy to show that the
AUC~\eqref{eq47} will then be given by
\begin{equation}\label{eq:3}
  AUC = \phi\left( \frac{a}{1+b^2} \right).
\end{equation}

\bigskip

Second, the normality assumption of the two distributions of the feature vector $X$ (the predictor)
has another set of interesting results and properties of the performance measures as well. E.g., the
early analysis by Fukunaga, in both mean and variance as a function of the training set size, is
fundamental \citep{Fukunaga1990Introduction, Fukunaga1989EffectsOf, Fukunaga1989EstimationOf}.

\bigskip

After the introduction above, it is very important to emphasize that it is not the intention of the
present article to discourage any practitioner from using the normal assumption for either the
feature vector or the scoring function. However, the objective is to illustrate how care and
prudence are needed when assuming normality as it may be violated severely even in very simple
setups, which may be counter-intuitive. The roadmap of the manuscript goes as follows.

\subsection{Manuscripat Roadmap}\label{sec:manuscripat-roadmap}
Section~\ref{sec:distr-score-funct} analyzes and derives mathematically, under the multinormal
assumption of the feature vector $X$, how the scoring function can be a single-tailed or
exponential-like distribution, which is a very sever violation from the binormal
model. Section~\ref{sec:properties-roc-auc} complements the article and provides some expected and
intuitive experimental results of the AUC under the multinormal assumption of the feature vector
$X$. This emphasizes the objective of the article that we do not discourage from using the normal
assumption; however, we urge practitioners for prudence towards adopting the
assumption. Section~\ref{sec:conclusion} concludes the article.


%% file: SecLoglikelihood.tex
\section{Distribution of Log-Likelihood Under Multinormal Assumption: a counter-intuitive
  example}\label{sec:distr-score-funct}
\begin{figure}[t]\centering
  \includegraphics[height=2.5in]{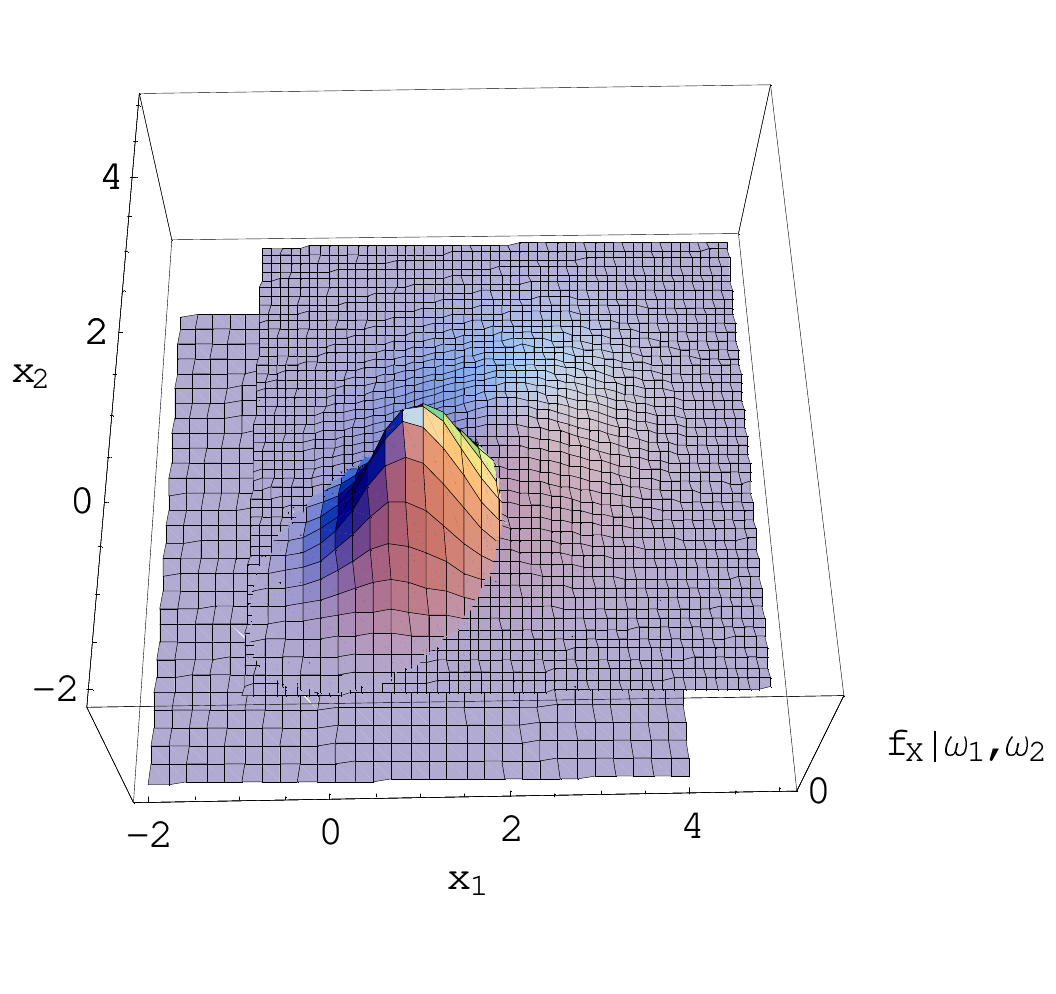}\hfil\includegraphics[height=2.5in]{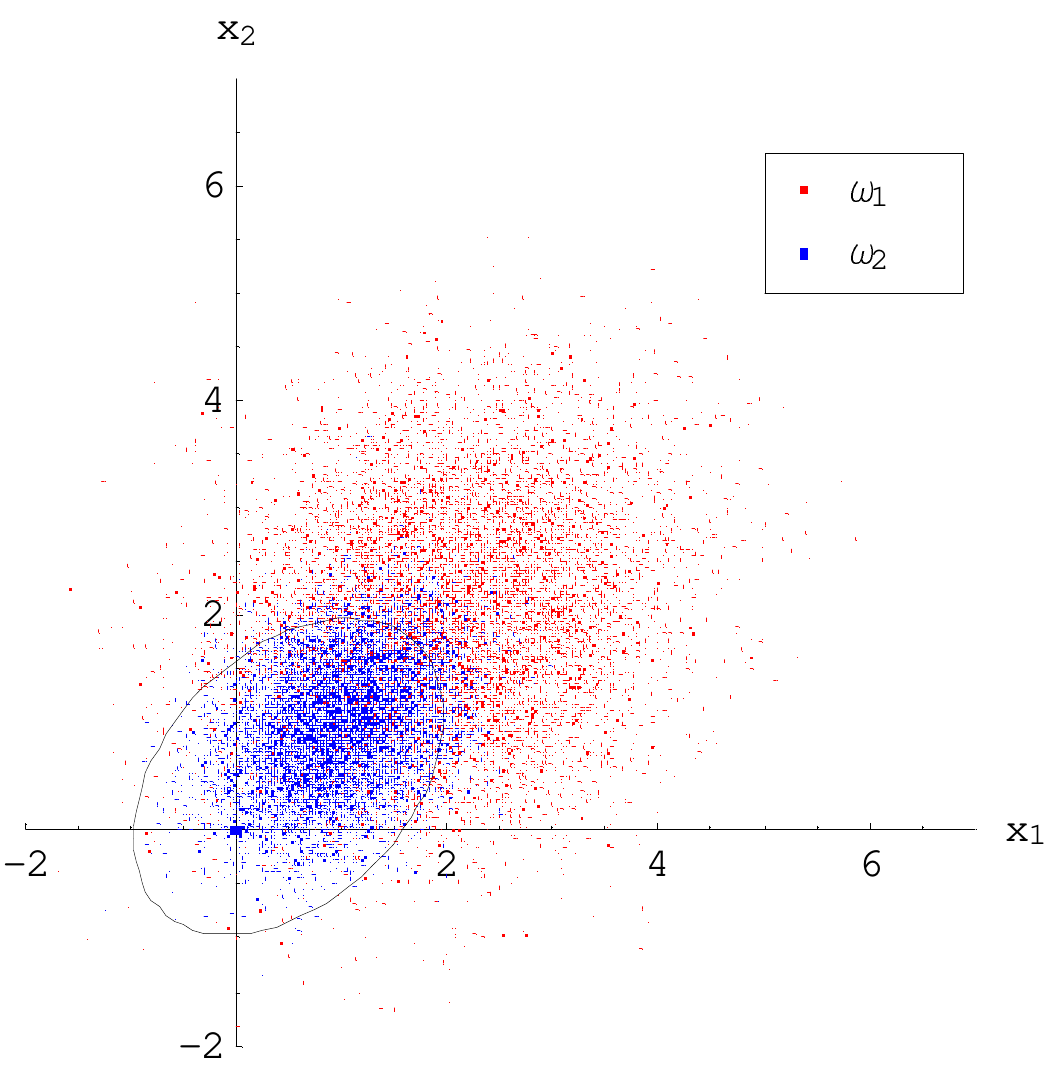}
  \caption{Two Binormal distributions. Left: a 3-D illustration of their Probability Density
    Function (PDF). Right: two simulated data sets from the two distributions, where the number of
    observations per class is $10,000$.}\label{Fig_3Dpdf}
\end{figure}

\begin{figure}[t]\centering
  \includegraphics[height=2.5in]{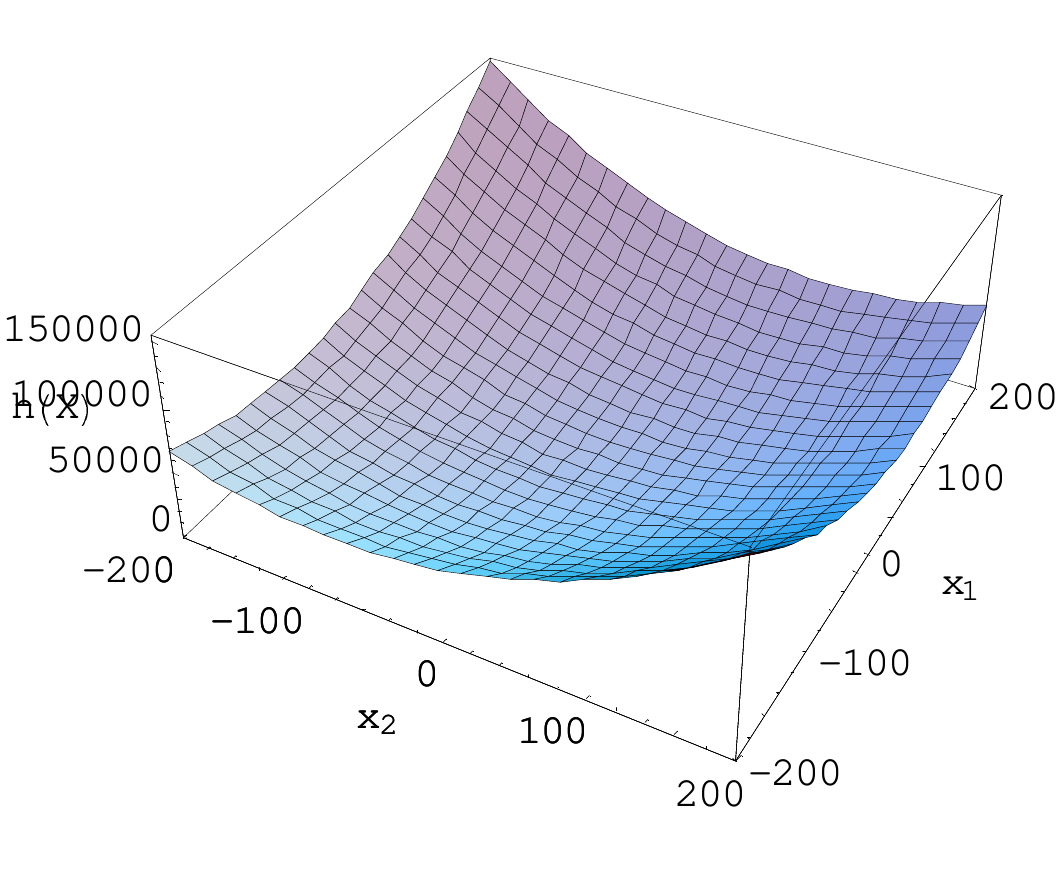}\hfill\includegraphics[height=2.5in]{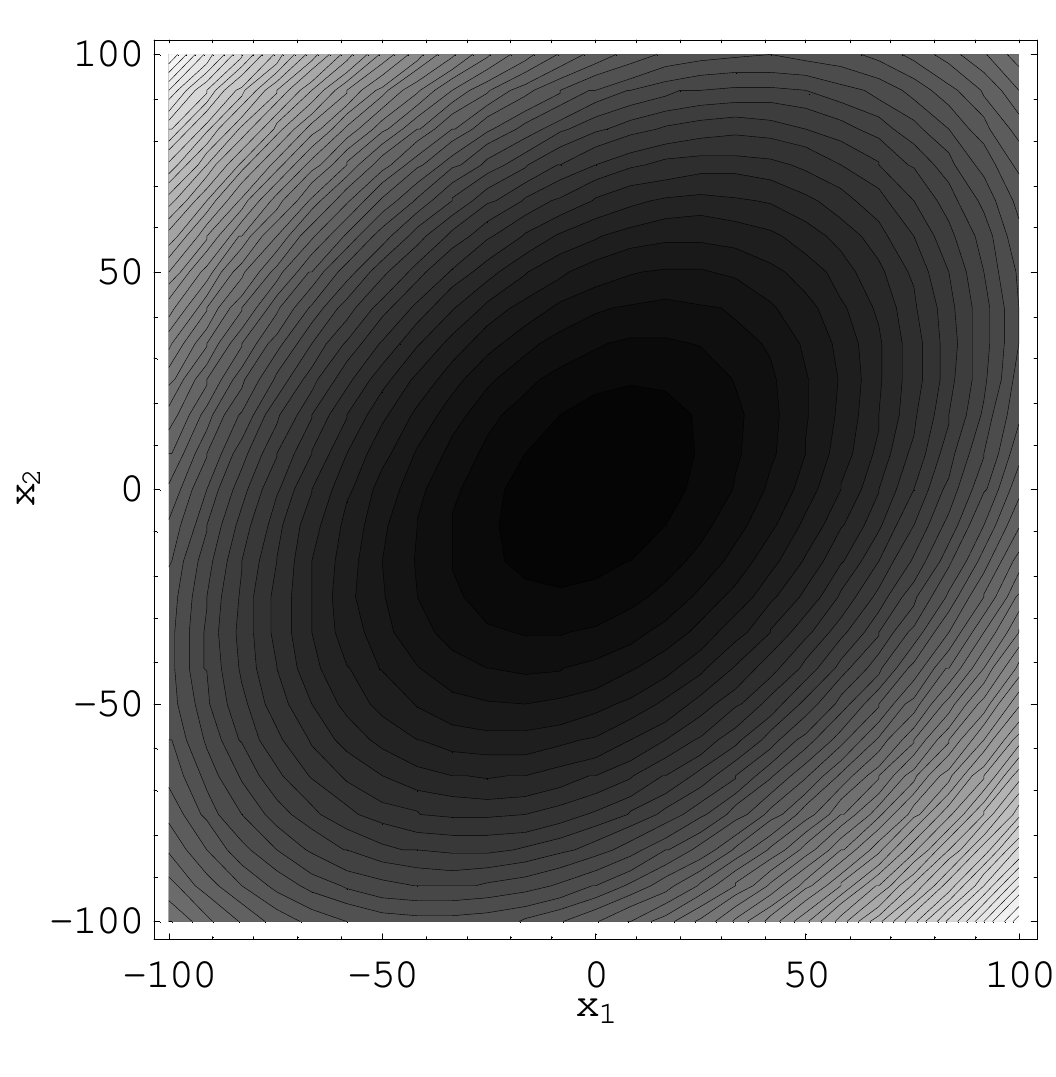}
  \caption{The log-likelihood ratio function of two features $x_{1}$ and $x_{2}$. Left: a 3-D
    representation. Right: its contour plot.}%
  \label{Fig_3DLR}
\end{figure}

It may be instructive to examine, analytically, what the log-likelihood ratio looks under the
popular data population of the feature vector:s the multinormal distribution. Consider two different
classes, $\omega_{1} $and $\omega_{2}$, whose $p$-dimensional feature vectors have the multinormal
distributions $F_{1}$ and $F_{2}$ respectively, described by the PDFs:
\begin{equation}
  f_{X}(x|\omega_{i})=\frac{1}{(2\pi)^{p/2}\left\vert {\Sigma_{i}}\right\vert^{1/2}}\exp\left[  {-\frac{1}{2}(x-\mu_{i}{)}^{\prime}\Sigma_{i}^{-1}(x-\mu_{i})}\right]  ,~i=1,2\label{eq107}
\end{equation}
The Bayes classifier is the optimal one; it has the minimum risk (see Section
\ref{sec:introduction}). The training for the Bayes classifier in the multinormal case requires only
the estimation of the mean vectors $\mu_{i}$ and the covariance matrices $\Sigma_{i}$. Assume that
the training set for $\omega_{1}$ is
$\mathbf{t}_{1}=\{t_{i}:t_{i}=(x_{i},\omega_{1})\},~i=1,\ldots ,n_{1}$, and the training set for
$\omega_{2}$ is $\mathbf{t}_{2}%
=\{t_{i}:t_{i}=(x_{i},\omega_{2})\},~i=1,\ldots,n_{2}$. The estimates of the population parameters are
given by \citep{Anderson2003AnIntroduction}:
\begin{align}
  \hat{\Sigma}_{i} &  =\frac{1}{n_{i}-1}\left[  {\sum\limits_{j=1}^{n_{i}}{(x_{j}-\hat{\mu}_{i})(x_{j}-\hat{\mu}_{i}{)}^{\prime}}}\right],\label{eq108}\\
  \hat{\mu}_{i} &  =\frac{1}{n_{i}}\sum\limits_{j=1}^{n_{i}}{x_{j}},~x_{j}\in\omega_{i}\label{eq:1}
\end{align}
The log-likelihood function (\ref{eq:Logliklyhood}), in combination with the estimated parameters
\eqref{eq108} and \eqref{eq:1}, assuming equal prevalence for the two classes and equal costs for
the two kinds of errors, can be written as:
\begin{equation}
  h(X)=-\frac{1}{2}\left[  {(X-\hat{\mu}_{1}{)}^{\prime}\hat{\Sigma}_{1}^{-1}(X-\hat{\mu}_{1})-(X-\hat{\mu}_{2}{)}^{\prime}\hat{\Sigma}_{2}^{-1}(X-\hat{\mu}_{2})}\right]  -\frac{1}{2}\ln\frac{\left\vert {\hat{\Sigma}_{1}}\right\vert }{\left\vert {\hat{\Sigma}_{2}}\right\vert }\label{eq109}
\end{equation}
It should be noted that if the a priori probabilities and costs, which form a particular threshold
value of the testing environment, are known they should be included in the log-likelihood
ratio. I.e., the log of the R.H.S. of inequality (\ref{eq14}) should be added to the R.H.S. of
(\ref{eq109}). In that case, the classifier is designed to be used in this environment having that
threshold. For demonstration, consider the population parameters to take the following values:%
\begin{align}
  \mu_{1} =\begin{pmatrix} 2\\ 2 \end{pmatrix},\quad \Sigma_{1}=\begin{pmatrix} 1 & .2\\ .2 & 1 \end{pmatrix},\quad \mu_{2} =\begin{pmatrix} 1\\ 1 \end{pmatrix},\quad \Sigma_{2}=\begin{pmatrix} .3 & .1\\ .1 & .3 \end{pmatrix}.\label{EQ_MuSigmaValues}
\end{align}
Consider that we train on a very large size of observations such that the estimated parameters are
almost the same as the true ones. Under these parameters the two PDFs (\ref{eq107}) of the two
classes are shown in \figurename~\ref{Fig_3Dpdf} (left). Two simulated data sets, one set for each
class with $10,000$ observations per class, are simulated from binormal distributions with the above
parameters and illustrated in \figurename~\ref{Fig_3Dpdf} (right).%
Under these parameter values, the log-likelihood ratio in (\ref{eq109}), along with its contour plot
is plotted in \figurename~\ref{Fig_3DLR}. The locus separating the two classes in
\figurename~\ref{Fig_3Dpdf} is obtained by solving $h(X)=0$ in (\ref{eq109}).%

\bigskip

We next find the probability distributions of $h(X)$ in (\ref{eq109}), under the two classes, which
is a function of the random vector $X=(x_{1},x_{2})^{\prime}$. Consider the transformation
$T:X\longrightarrow(h,d)$, where $d$ is a dummy variable set as%
\begin{equation}
  d=x_{1} \label{EQ_dummyVar}%
\end{equation}
The dummy variable here is introduced just for clarity and we could have written
$T:X\longrightarrow(h,x_{1})$. We shall, no longer, refer to $d$; rather, we refer to its genuine
$x_{1}$. This transformation is $1:2$, which produces two values of the new vector $(h,x_{1})$ at
every value of the vector $X$. In other words, solving (\ref{eq109}) and (\ref{EQ_dummyVar}) for
$x_{1}$ and $x_{2}$ gives two solutions. By adding these two solutions, and calculating the Jacobian
of the transformation it can be shown that the joint PDF of $h(X)$ and $x_{1}$, for $X\sim F_{1}$,
is given by:%
\begin{equation}
  f(h,x_{1}|\omega_{1})=\exp\left[-.385h-.074x_{1}^{2}+1.805x_{1}-1.243\sqrt {r}\right] \times  \frac{.00157}{\sqrt{|r_{1}|}}\left(\exp\left[.178x_{1}\sqrt{r}\right]+\exp\left[2.49\sqrt{r}-.178x_{1}\sqrt{r}\right]\right)\label{EQ_pdf_h_x1_w1}
\end{equation}
and, for $X\sim F_{2}$, is given by:
\begin{equation}
  f(h,x_{1}|\omega_{2})=\exp\left[-1.385h-.074x_{1}^{2}+1.805x_{1}-1.243\sqrt{r}\right]\times\frac{.00157}{\sqrt{|r_{1}|}}\left(\exp\left[.178x_{1}\sqrt{r}\right]+\exp\left[2.49\sqrt {r}-.178x_{1}\sqrt{r}\right]\right),\label{EQ_pdf_h_x1_w2}
\end{equation}
where%
\begin{equation}
  r_{1}=1.91+.866h+x_{1}-x_{1}^{2} \label{EQ_parab1}%
\end{equation}

The conditional joint density functions (\ref{EQ_pdf_h_x1_w1}) and (\ref{EQ_pdf_h_x1_w2}) are defined only
on a parabolic area determined in the $h$-$x_{1}$ space by the parabola $r_{1}$ in
(\ref{EQ_parab1}). In general, under different values of the mean vectors and covariance matrices
(\ref{EQ_MuSigmaValues}) this area is a general conic section.

\begin{figure}[t]\centering
  \includegraphics[height=2.5in]{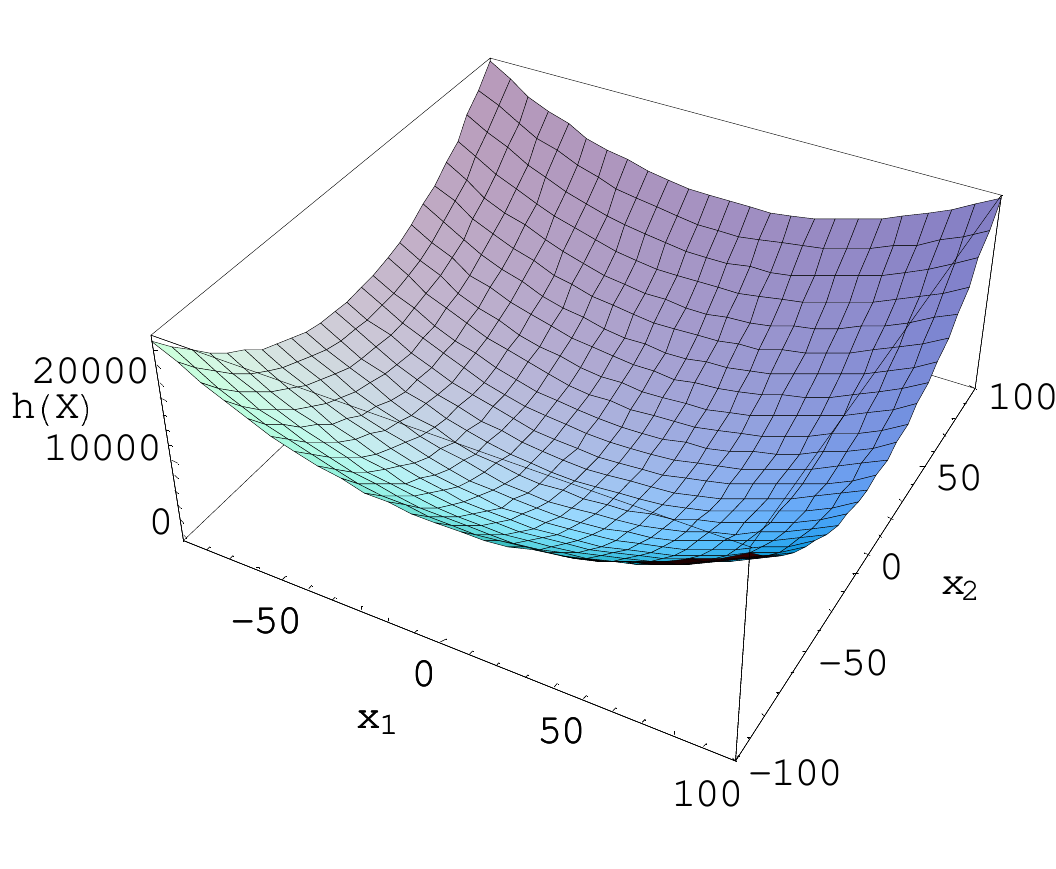}\hfill\includegraphics[height=2.5in]{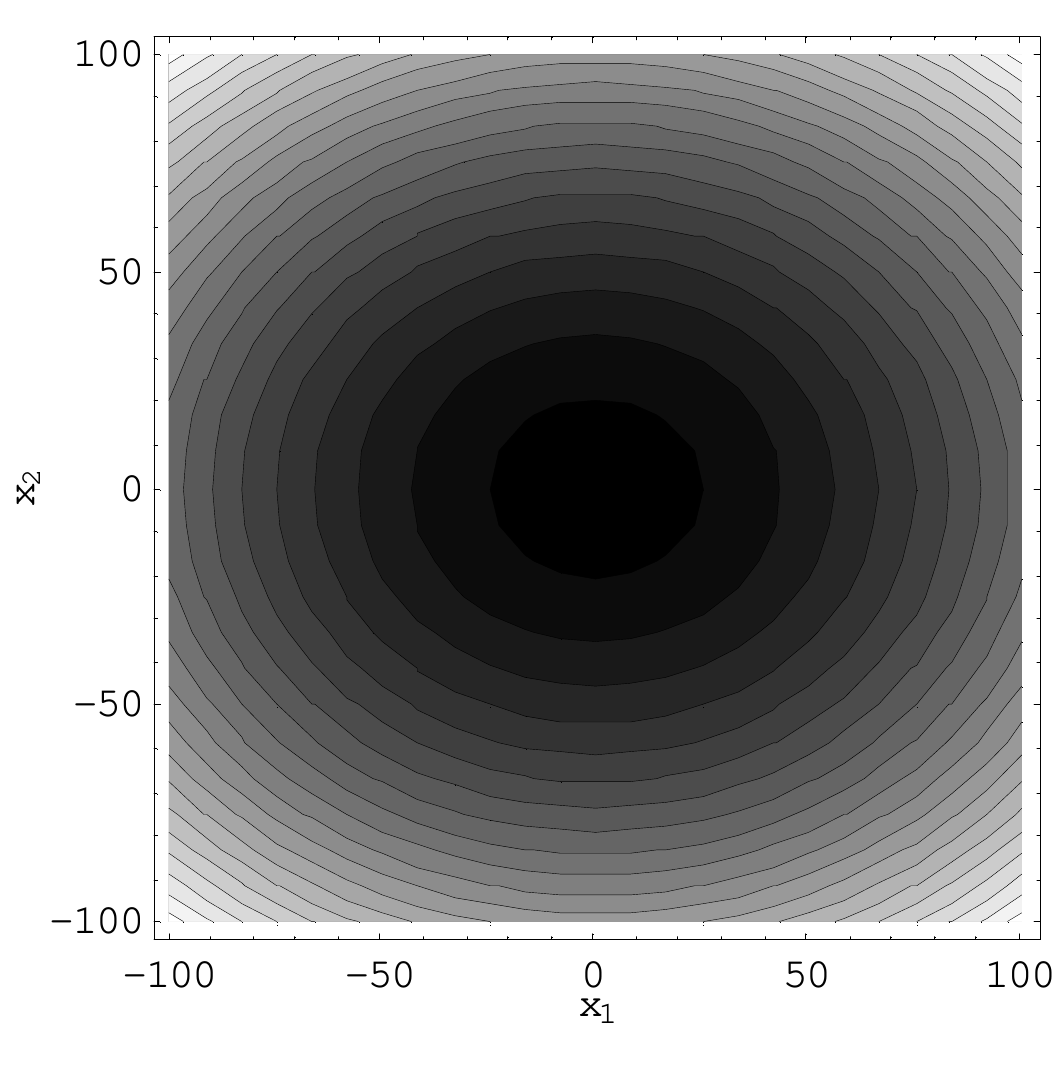}
  \caption{The log-likelihood ratio function of two features $x_{1}$ and $x_{2}$ after simultaneous
    diagonalization for the two covariance matrices $\Sigma_{1}$ and $\Sigma_{2}$. Left: a 3-D
    representation. Right: its contour plot.}\label{Fig_Diag3DLR}
\end{figure}

For completeness, and a little bit off topic, the following should be mentioned. One could have
simultaneously diagonalized the matrices $\Sigma_{1}$ and $\Sigma_{2}$---said differently,
transformed $x_{1}$ and $x_{2}$ to $x_{1}^{\prime}$ and $x_{2}^{\prime}$ such that the new variables
have diagonalized covariance matrices---to get rid of the cross terms in the new space of
$h$-$x_{1}^{\prime }$. For the topic of simultaneous diagonalization the reader may be referred to
\cite{Fukunaga1990Introduction} or for more rigorous analysis to \cite{Schott2005Matrix} or
\cite{Searle1982Matrix}. After performing the simultaneous diagonalization to the matrices
$\Sigma_{1}$ and $\Sigma_{2}$ and proceeding as described above, the joint density function of $h$
and $x_{1}^{\prime}$ can be shown to be
\begin{subequations}
  \begin{align}
    f_{H,X_{1}^{\prime}}(h,x_{1}^{\prime}|\omega_{1}) &=\frac{.00232\exp[-.33h-.166x_{1}^{\prime2}+2.15x_{1}^{\prime}]}{\sqrt{|r_{2}|}},\label{EQ_pdf_h_x1_w1_Diag}\\
    f_{H,X_{1}^{\prime}}(h,x_{1}^{\prime}|\omega_{2}) &=\frac{.00232\exp[-1.33h-.166x_{1}^{\prime2}+2.15x_{1}^{\prime}]}{\sqrt{|r_{2}|}},\label{EQ_pdf_h_x1_w2_Diag}\\
    r_{2} &=2.08+h+1.29x_{1}^{\prime}-x_{1}^{\prime2},\label{EQ_parab2}
  \end{align}
\end{subequations}
where $r_{2}$ is the region on which the joint density of $h$-$x_{1}^{\prime}$ is defined. The
log-likelihood ratio (\ref{eq109}), after simultaneous diagonalization, is illustrated in 3-D, as
well as, contour plot in \figurename~\ref{Fig_Diag3DLR}. The virtue of simultaneous diagonalization
is obvious from Equations \eqref{EQ_pdf_h_x1_w1_Diag}--\eqref{EQ_parab2}.%

\bigskip

\begin{figure}[t]\centering
  \includegraphics[height=1.4in]{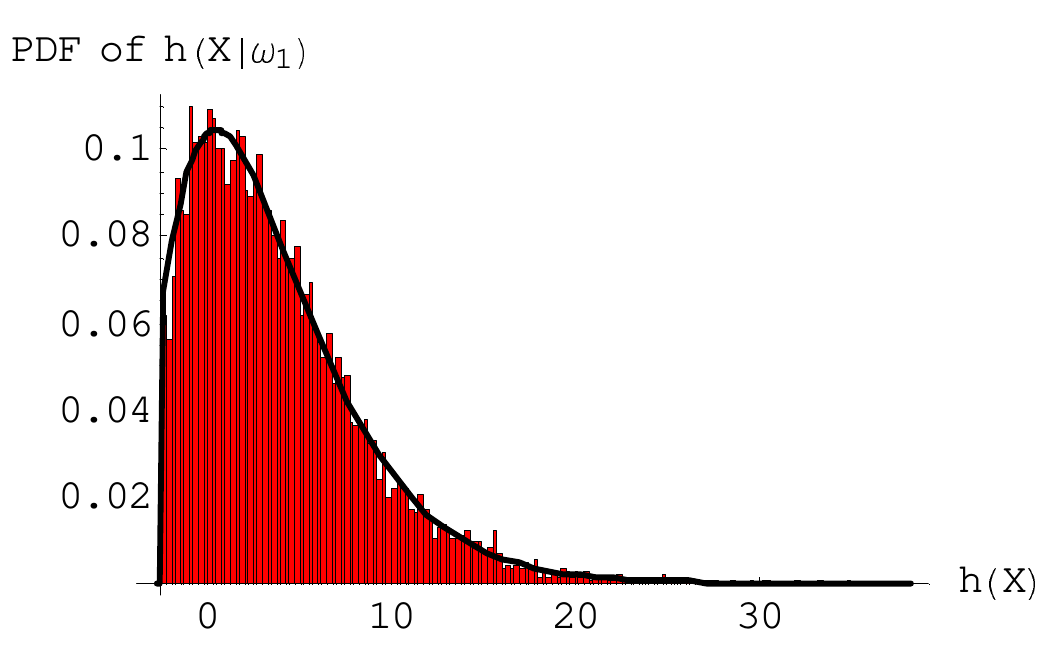}\hfill\includegraphics[height=1.4in]{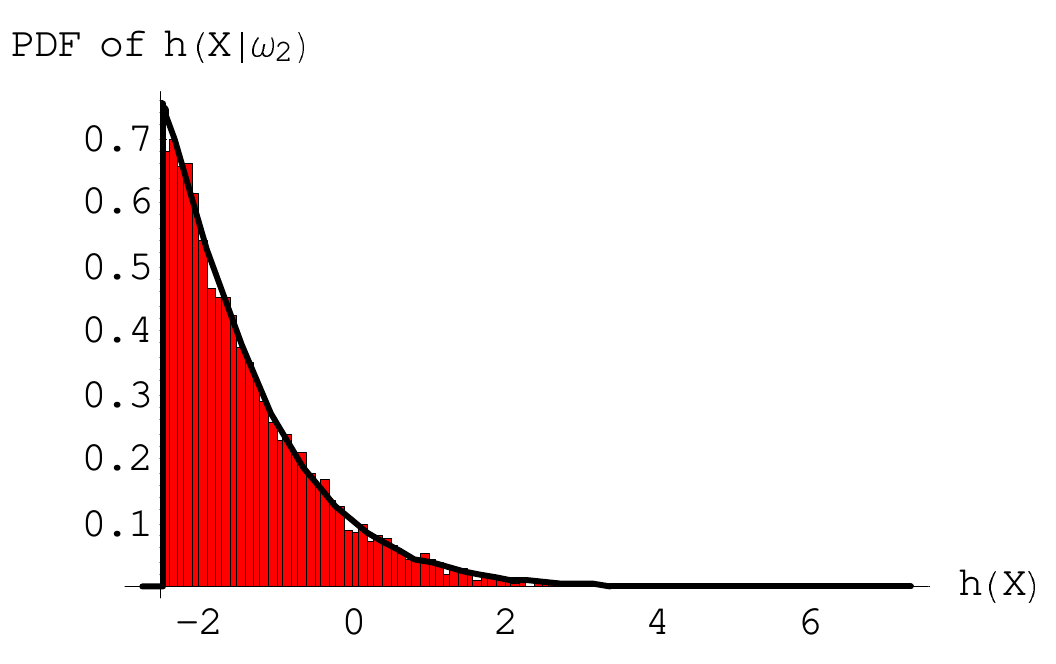}\hfill\includegraphics[height=1.4in]{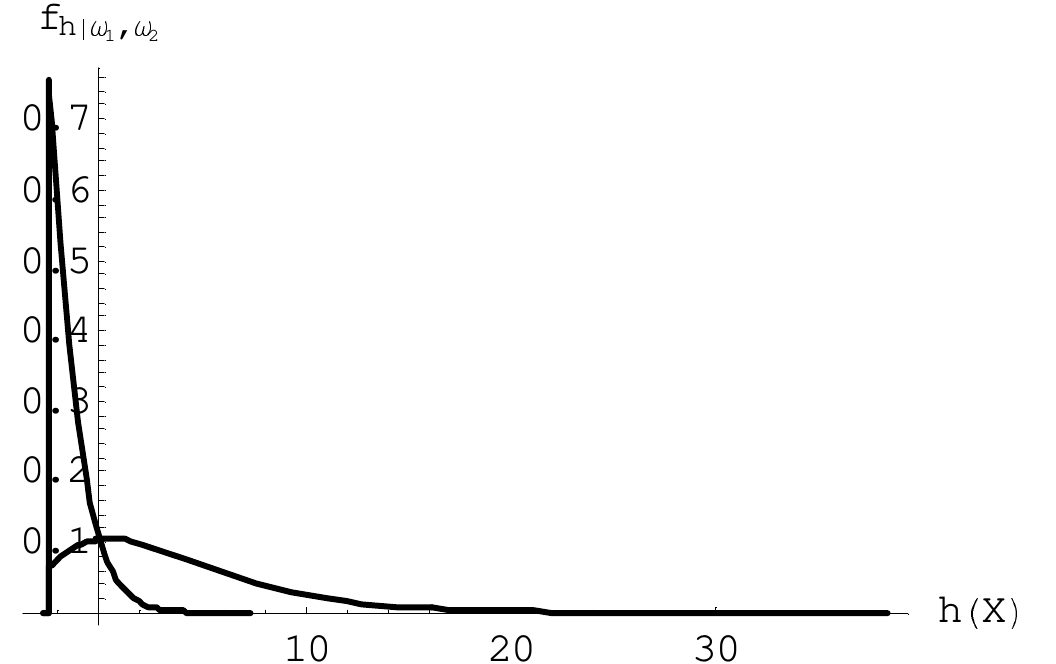} \caption{The
    PDF of the log-likelihood ratio under $\omega_{1}$ (left) and $\omega_2$ (middle) obtained from
    mathematical analysis, along with its histogram obtained from a simulation study. The same two
    PDFs are plotted together on the same scale (right)}\label{Fig_LRpdf1}
\end{figure}

Back to \eqref{EQ_pdf_h_x1_w1}--\eqref{EQ_pdf_h_x1_w2}, unfortunately, a closed form integration for
(\ref{EQ_pdf_h_x1_w1}) and (\ref{EQ_pdf_h_x1_w2}) over $x_{1}$, to obtain the marginal probabilities
$f(h|\omega_{1})$ and $f(h|\omega_{2})$, is not available. However, we always can obtain a numerical
solution to the problem by carrying out the integration over $x_{1}$ for every desired value of
$h$. The marginal PDFs of $h$, conditional on $\omega_{1}$ and $\omega_{2}$, are obtained by the
described technique and illustrated in \figurename s~\ref{Fig_LRpdf1} (the two left figures). In
addition, these two figures show the histograms of $h$ obtained from simulating testing observations
from the distributions $F_{1}$ and $F_{2}$ and obtaining the log-likelihood ratio $h$ for every
observation. The figures show how well both, the histogram and the mathematical solution, are highly
consistent. \figurename~\ref{Fig_LRpdf1} (right) shows the two PDFs, together on the same scale, for
the classification purpose.%

\bigskip

It is extremely important to comment on the result illustrated in
\figurename~\ref{Fig_LRpdf1}. Although the data are coming from binormal distributions, the
log-likelihood ratio is not distributed as normal distribution; both are single-tailed. Moreover,
$f_{h|\omega_{2}}$, in no way, can be approximated to a normal distribution. It has an abrupt
behavior that makes it resemble more the exponential distribution. This simple example provides an
important caveat to the exaggerated use of normality for the log-likelihood ratio.


%% file: SecPerofrmance.tex
\section{Properties of AUC Under Multinormal Assumption: expected
  behavior}\label{sec:properties-roc-auc}
\begin{figure}[t]\centering
  \includegraphics[height=2.3in]{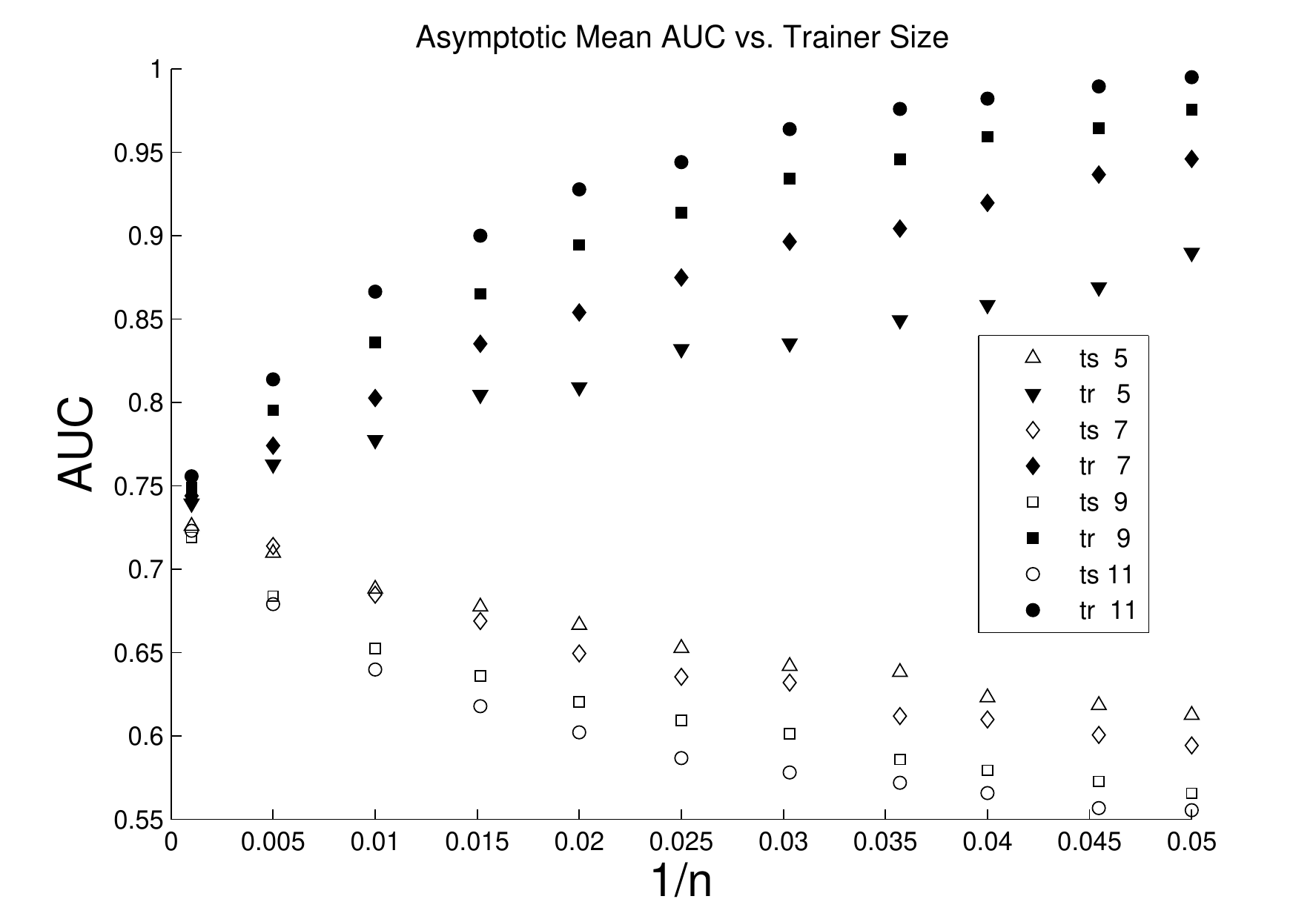}\hfil\includegraphics[height=2.3in]{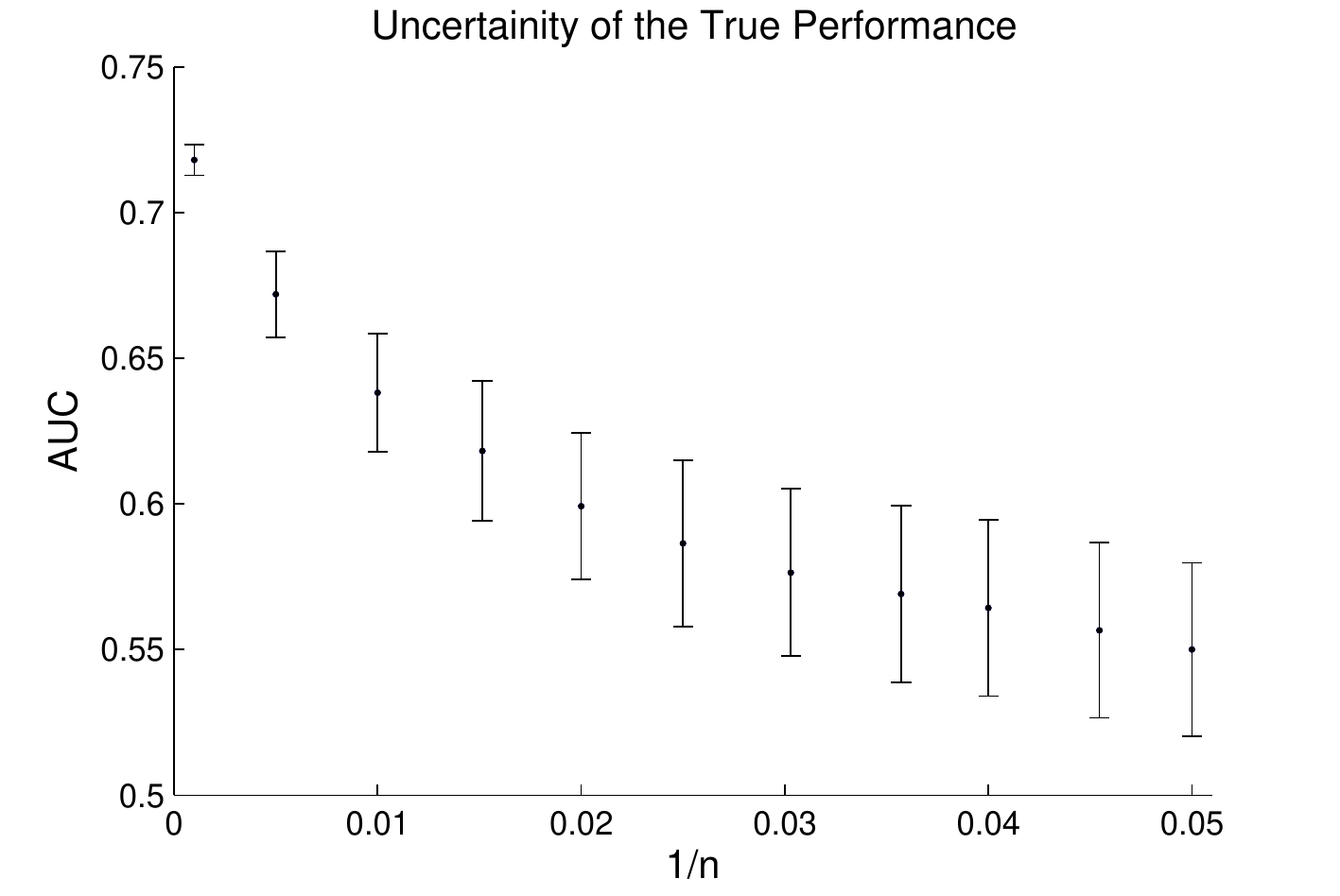}
  \caption{Classifier performance. Left: mean AUC of the Bayes classifier.  For every training
    sample size $n$, the classifier is tested on pseudo-infinite testers (represented as ``ts'') and
    tested as well on the same training sample ( represented as ``tr''). Each curve shows the
    average performance over 100 MC trials. The numbers in the legend are the dimensionalities of
    the feature vectors. Right: uncertainty (variance) around the mean performance of the Bayes
    classifier, for 11 features, vs. the size of the training data set. Asymptotically, the
    variability vanishes.}\label{fig3IEEE}
\end{figure}

This section contrasts Section~\ref{sec:distr-score-funct}. We present some of the expected behavior
under the multinormal assumption of the feature vector. To exhibit the basic structure of the
problem under the practical limitation of a finite-training set, we carried out simulations inspired
by \cite{Chan1999ClassifierDesign} and the work of \cite{Fukunaga1989EffectsOf,
  Fukunaga1989EstimationOf}. In our simulation, we assume that the feature vector has the
multinormal distribution with the following parameters: $\mu_{1}={0}$, $\mu_{2}=c\mathbf{1}$, and
$\Sigma_{1}=\Sigma_{2}=\mathbf{I}$, where $\mathbf{0}$ and $\mathbf{1}$ are the vectors all of whose
components are zeros and ones respectively, $\mathbf{I}$ is the identity matrix, and $c$ is a
constant used for adjusting the separation between the two classes. A fundamental measure for that
is the Mahalanobis distance between the mean vectors; it is defined as:%
\begin{equation}
  \Delta=\left[{(\mu_{1}-\mu_{2}{)}^{\prime}\Sigma^{-1}(\mu_{1}-\mu_{2})}\right]^{1/2}.\label{eq11IEEE}
\end{equation}
It expresses how these two vectors are separated from each other with respect to the spread
$\Sigma$. In the simulation of the present example, the Mahalanobis distance is $c^{2}p$. In this
simulation, illustrated in \figurename~\ref{fig3IEEE}, the value $c$ is adjusted for every
dimensionality to obtain the same asymptotic AUC that can be obtained form~\eqref{eq:3}.
This allows us to isolate the effect of the variation in training set sizes.  Typically, the
simulations described in this context used $\Delta=0.8$. For the time being, it is assumed that
$n_{1}=n_{2}=n$, which is referred to as the training set size per class. For a particular
dimensionality, and for particular data set size $n$, two training data sets are generated using the
above parameters and distributions. When the classifier is trained, it will be tested on a
pseudo-infinite test set, here 1000 cases per class, to obtain a very good approximation to the true
AUC for the classifier trained on this very training data set; this is called a single realization
or a Monte-Carlo (MC) trial. Many realizations of the training data sets with same $n$ are generated
over MC simulation to study the mean and variance of the AUC for the Bayes classifier under this
training set size. The number of MC trials used is 100.

\bigskip

Since any classifier is designed using a finite-size data set, its true performance---the
performance obtained from expectation over the population---is dominated by the size of this
set---assuming fixing the distribution of the data. When the classifier is re-designed using
different training set size, the expected performance will vary. This is because the limited size
training set has some, not all, of the information represented in the
population. \figurename~\ref{fig3IEEE} (left) illustrates this for the set of experiment parameters
explained above, where it plots the mean AUC vs the reciprocal of the training set size under
different dimensionality $p$. On the other hand, if the classifier is re-designed, using another
data set having the same size, the performance will vary as well. This is because the performance
measure is a function of the training data set; hence it is a random
variable. \figurename~\ref{fig3IEEE} (right) is an illustration of this fact for the special case of
$p=11$, and $c=0.27$.

\bigskip

Several important observations can be made from these results. As was expected, for training size
$n$ the mean apparent AUC, i.e., coming from testing on the same training data set, is upwardly
biased from the true AUC.  It should be cautioned that this is on the average, i.e., over the
population of all training sets; it is possible that for a single data set (single realization) the
apparent performance can be better or worse than the true one. In addition, the classifier had the
same asymptotic performance, approximately 0.74, for all dimensionalities in the simulation (by
construction as explained above).


%% file: SecConclusion.tex
\section{Conclusion}\label{sec:conclusion}
First, this article introduced the binary classification problem and provided a brief account of its
mathematical foundation. In this introduction we distinguished between the distribution of the input
feature vector (the predictor) and the distribution of the output scoring function that will be
compared to a threshold value to provide the final class prediction (the response). The distribution
of the output scoring function determines the properties of the performance measure of the
classifier, including the error rate, the Receiver Operating Characteristic (ROC) curve, and the
Area Under the Curve (AUC), among many other measures. There is a well established literature, both
theoretically and experimentally, for the performance of classification rules under the normality
assumption of either the input feature vector or the output scoring function. This rich literature
makes it very tempting and, in many cases very beneficial, to adopt either of these two normality
assumptions. Second, the article proved with a counter example how this normality assumption of the
scoring function may be severely violated under a very simple setup of normal assumption of the
input feature vector. However, we frequently see the Central Limit Theorem (CLT) at work in higher
dimensions driving the final scoring function towards the binormal assumption and the ROC curve
towards the double-normal-deviate plot. Third, the article provided the complementary balancing
message and illustrated experimentally some of the expected properties of the AUC under the
multinormal assumption of the input feature vector. To recap, the article does not discourage the
practitioners from using the normality assumption; however, the article calls for prudence when
adopting it.


%% file: SecAcknowledgment.tex
\section{Acknoledgment}\label{sec:acknoledgment}
The author is grateful to the U.S. Food and Drug Administration (FDA) for funding an earlier stage
of this project. Special thanks and gratitude, in his memorial, to Dr. Robert F. Wagner the
supervisor and teacher, or Bob Wagner the big brother and friend. He reviewed a very early version
of this manuscript before his passing away.
